\documentclass[letterpaper, 10 pt, conference]{ieeeconf}  

\IEEEoverridecommandlockouts                              

\usepackage{graphics} 
\usepackage{epsfig} 
\usepackage{mathptmx} 
\usepackage{times} 
\usepackage{amsmath} 
\usepackage{amssymb}  

\title{\LARGE \bf CircuitBot: Learning to Survive with Robotic Circuit Drawing\\

\author{Xianglong Tan$^{1,2}$, Weijie Lyu$^{2}$ and Andre Rosendo$^{2}$
\thanks{This work was supported by the National Natural Science Foundation of China, project number 61850410527, and the Shanghai Young Oriental Scholars project number 0830000081.}
\thanks{$^{1}$Department of Physics,
        University of Cambridge, UK
        }%
\thanks{$^{2}$Living Machines Laboratory, School of Information Science and Technology, ShanghaiTech University, China
        }%
}
}

\begin{document}

\maketitle
\thispagestyle{empty}
\pagestyle{empty}

\begin{abstract}

Robots with the ability to actively acquire power from surroundings will be greatly beneficial for long-term autonomy, and to survive in dynamic, uncertain environments. In this work, a scenario is presented where a robot has limited energy, and the only way to survive is to access the energy from a power source. With no cables or wires available, the robot learns to construct an electrical path and avoid potential obstacles during the connection. We present this robot, capable of drawing connected circuit patterns with graphene-based conductive ink. A state-of-the-art Mix-Variable Bayesian Optimization is adopted to optimize the placement of conductive shapes to maximize the power this robot receives. Our results show that, within a small number of trials, the robot learns to build parallel circuits to maximize the voltage received and avoid obstacles which steal energy from the robot.

\end{abstract}

\section{INTRODUCTION}

With recent development in the field of robotic hardware, sensing and machine learning, we have witnessed great progress of robotic applications in 
industrial settings. \cite{8793956}, \cite{duckett2018agricultural}, \cite{Birrell2020AFR}. The advance in robotics naturally raises the demand for fully-autonomous, long-lived, and self-evolving robots \cite{10.1371/journal.pone.0186107}, keeping human effort further out of the loop. As with living species, robots consume energy to perform different tasks. Industrial robots deployed in the real world are typically powered by uninterruptible power supply without the fear of running out of energy. For mobile robots that are powered by batteries, however, current energy storage technologies suffer from low gravimetric and volumetric energy densities, significantly limiting the mobility, operational times, and performance of the robots \cite{Yangeaar7650}. This stark outlook necessitates a new energy acquisition ability for robots.

Previous researches on robotic self-powered ability focus on the design of power system. Solar panels are widely-used for robots to harvest energy from ambient sources to recharge batteries. Solar panels are normally low-cost and light-weighted but suffer from low efficiency and restrictive application environment. In a series of works on energy harvesting \cite{Xu2010}, \cite{Wang102}, \cite{Pan1947}, researchers demonstrate the powering of nanobots from ambient mechanical energy. In a more recent work \cite{doi:10.1021/acsenergylett.9b02661}, a novel robotic power system is built powered by scavenging energy from external metals. These works provide significant progress and insights on how robots can survive through energy in the environments. 

This paper presents a novel approach for robots to leverage ambient electrical power sources to survive in resource-limited, uncertain environments. Instead of using wires or cables, the robot constructs its own electrical path to connect a power supply with graphene-based conductive ink. The flexibility of the conductive ink enables various circuit patterns to avoid potential obstacles during the path. We adopt a novel Bayesian Optimization algorithm to optimize the circuit's patterns to maximize the power robot received. This paper shows that robots can use Bayesian learning techniques to optimize their own self-drawn electric circuits in very few trials when put in a condition where their survival is at stake.


\section{ROBOTIC CIRCUIT DRAWING}

\subsection{Conductive Ink}

Graphene-based conductive ink has shown great potential in printing flexible electronics for its low cost, high connectivity and can be applied directly on materials like textile\cite{C7TC03669H}, paper, and other diverse flexible substrates\cite{Huang:2011aa}. Compared to metal-based conductive ink, it is low-toxic, environment friendly, and easy to make and store \cite{Pan_2018}. In this work, we follow the instructions in \cite{Saidina_2019} to fabricate customized graphene-based conductive ink. The ink is made of $5$ $wt\%$ graphene flakes, $0.5$ $wt\%$ graphene dispersion, and $94.5$ $wt\%$ water, which has a sheet resistance of approximately $2$ $\Omega/sq$.

\subsection{Experimental Setup}
\begin{figure}[t!]
\centerline{\includegraphics[width=0.48\textwidth]{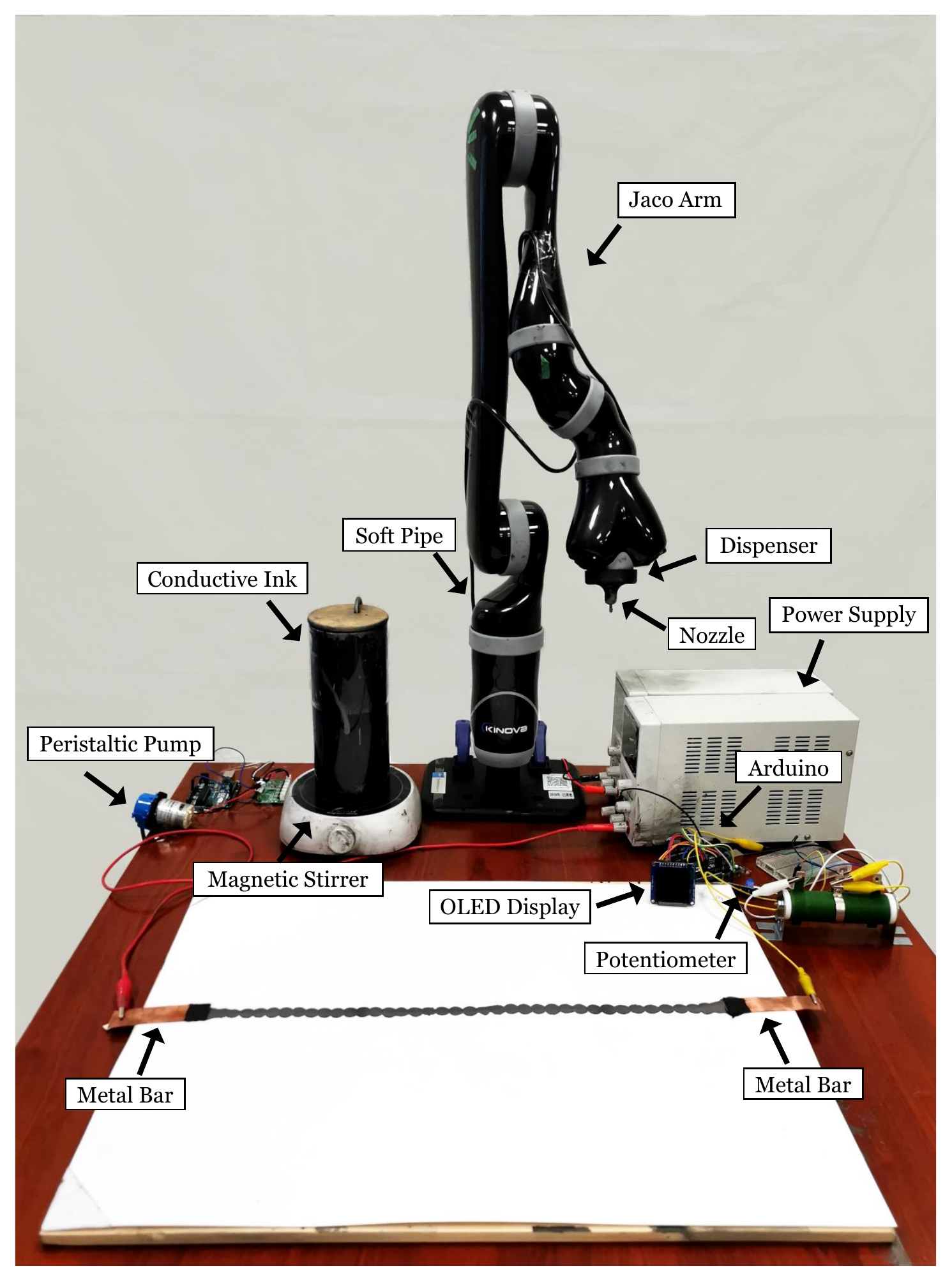}}
\caption{Experimental setup of the circuit drawing robot. The Kinova 6DOF Jaco Arm first moves to an initial Cartesian coordinate with the end-effector (nozzle) at $5~cm$ from the paper. The ROS controller sends instructions to both the arm and an Arduino to begin the circuit drawing. The circuit drawing movements are encoded as a list of Cartesian coordinates and sent to the arm. The Arudino then receives the state of the arm through ROS and sets the on/off of the peristaltic pump to control the ink flow. The connection starts to show conductivity after the ink dries (30 minutes). The output voltage achieved is then shown on the OLED display.}
\label{Exp_setup}
\end{figure}

\begin{figure}[htbp]
\centerline{\includegraphics[width=0.4\textwidth]{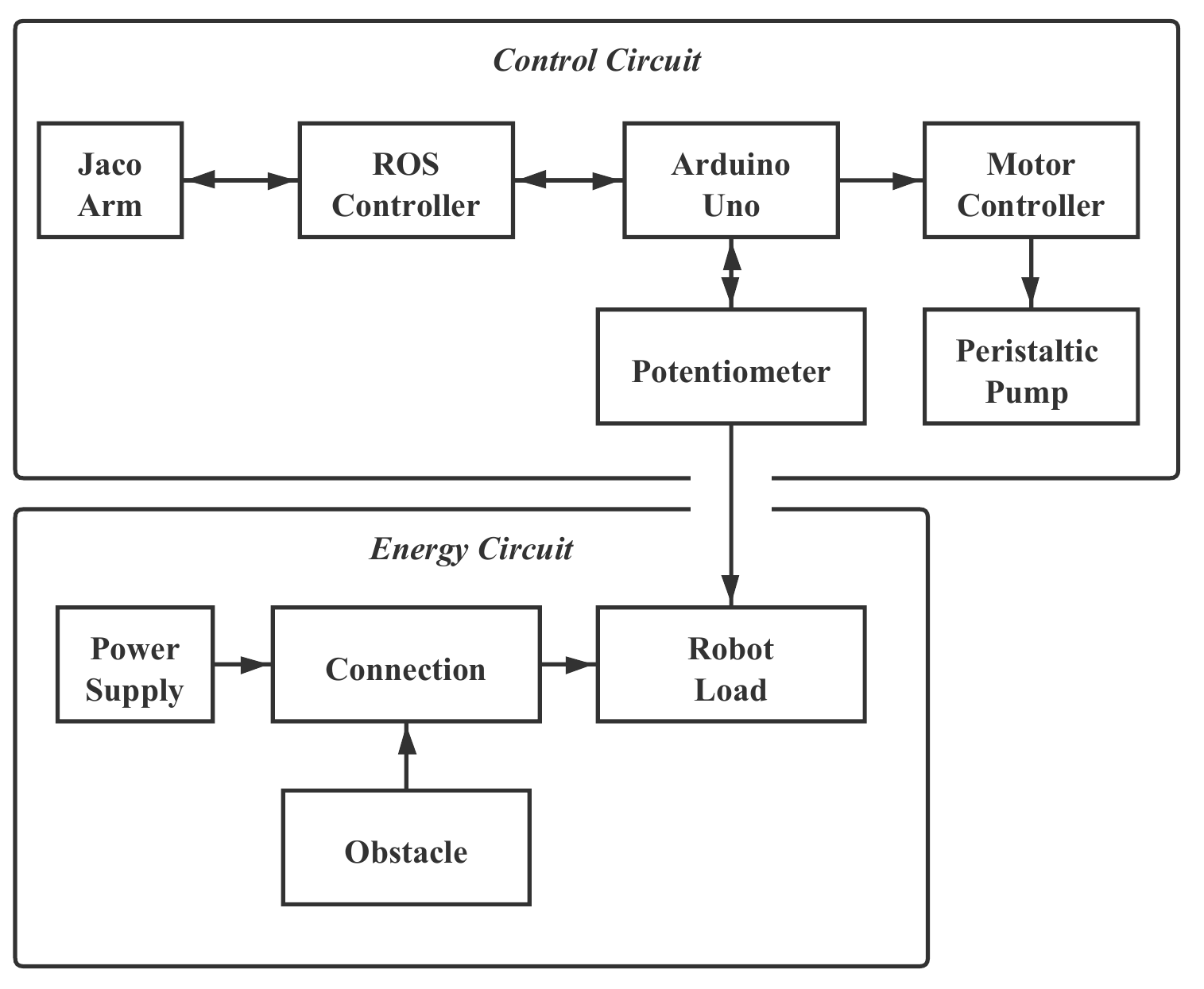}}
\caption{Electrical and control architectures.}
\label{Exp_hardwareBlock}
\end{figure}

The experimental setup is shown in Fig.~\ref{Exp_setup}. The conductive ink is prepared and stored in a glass jar placed on a magnetic stirrer to prevent the graphene from solidification. The glass jar is linked to a soft pipe which is connected to a peristaltic pump. The pump pushes the ink towards a nozzle which is held by a 3D-printed dispenser at the end-effector of the Kinova 6DOF Jaco Arm. An Arduino Uno is used to control the speed of ink flow.
The robot arm is controlled by Moveit (https://github.com/ros-planning/moveit). Two metal bars are fixed on the cardboard, representing the terminals of the robot and the power source. The circuit block diagram of the robotic system is shown in
Fig.~\ref{Exp_hardwareBlock}.


\section{CIRCUIT OPTIMIZATION}

\subsection{Motivation}

The conductivity of the connection between the robot and the power source is a major concern. Connection with low resistance would provide lower loss and higher current when powering the robot.
The conductivity of the ink can be enhanced by increasing the concentration of the graphene flakes. However, adding more graphene would sacrifice the ink's flowability, resulting in the soft pipe being jammed by the ink.
Another method is to increase the speed of the ink flow. However, the cardboard on which the robot draws has a very limited capacitor for water absorption. With too much liquid on the surface, the patterns are unstable and easy to spread; meanwhile, the dry time for the ink to show conductivity increases significantly.
The issue motivates us to think from an electrical point of view. The approach we investigate in this work is to create parallel electrical paths to reduce the resistance of the connection. In this paper, we refer to this task as circuit optimization.

\subsection{Drawing Procedure}
\begin{figure}[ht]
\centerline{\includegraphics[width=0.48\textwidth]{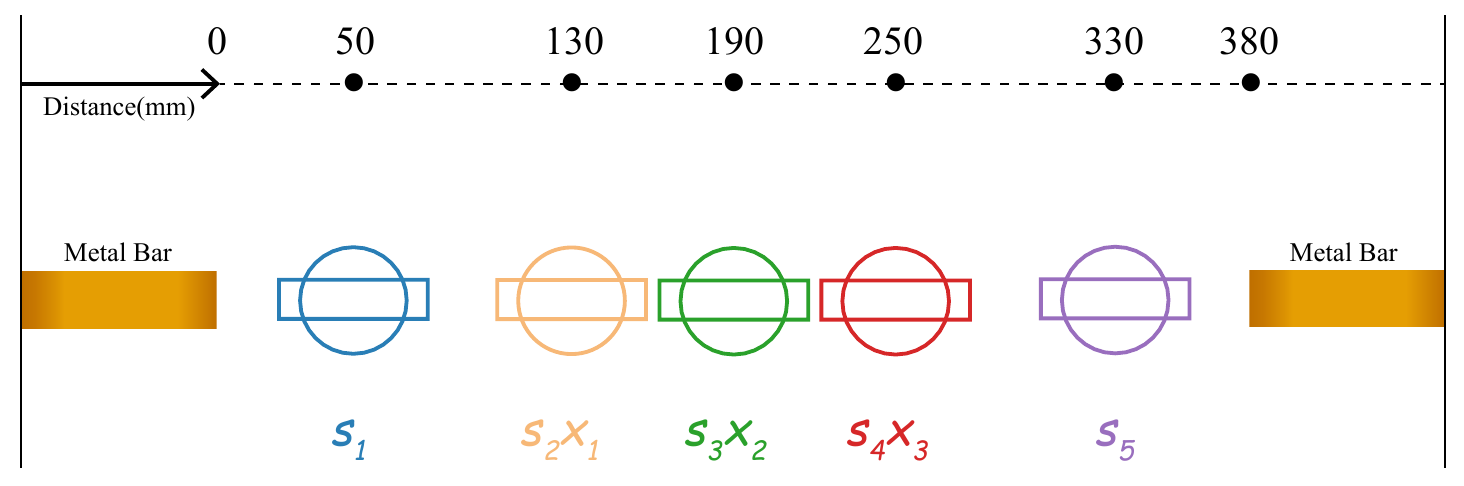}}
\caption{Two types of circuit shapes (i.e., line and circle) are selected to form the connection between two metal bars. The  circle  diameter  and  line length are both $100$ $mm$. In total five shapes $s_1-s_5$ are drawn sequentially on a $380\times100$ $mm^2$ workspace. The central coordinates of $s_1$ and $s_5$ are fixed. The central coordinates of $s_2-s_4$ are denoted as $x_1-x_3$, which have a horizontal displacement range of $\pm20~mm$.}
\label{Drawing_procedure}
\end{figure}

The circuit drawing procedure is illustrated in Fig.~\ref{Drawing_procedure}, and a list of control parameters was identified, as shown in Table~\ref{parameter_describe}. Two metal bars are placed on the cardboard with a distance of $380$ $mm$. One is connected to a $30~V$ DC power supply; the other one is connected to a resistive load ($45\Omega$). A total of five shapes are drawn sequentially to connect the two metal bars. $s_1-s_5$ denotes the five shapes drawn from the left metal bar to the right. $x_1-x_3$ are the center coordinates of $s_2-s_4$. The voltage of the load is measured by a potentiometer as an indicator of the quality of the connection. We selected two categories of shapes, line and circle, to represent series and parallel circuit patterns. For value of $s_1-s_5$, 0 (zero) denotes line and 1 (one) denotes circle. The circle diameter and line length are both set to be $100$ $mm$. The workspace has a size of $380\times100$ $mm^2$, the centers of $s_1$ and $s_5$ are fixed. The initial center coordinates of $s_2-s_4$ are evenly distributed across the workspace and can displace horizontally for $\pm20~mm$. All the drawn circuits are dried for $30$ $mins$ before measurement, and ink flow speed is kept constant.

\begin{table}[htbp]
\caption{Input Variables}
\begin{center}
\begin{tabular}{|c|c|c|c|c|}
\hline
Variable & Name & Quantity & Type & Value \\ \hline
$s$ & Shape Indicator & 5 & Categorical & 0, 1 \\ \hline
$x$ & Central Coordinate & 3 & Continuous & (-20, 20) \\ \hline
\end{tabular}
\label{parameter_describe}
\end{center}
\end{table}

\subsection{Reliability of Circuit Drawing}

An investigation was conducted to examine the reliability of lines and circles circuits. For each type of shape 20 trials were conducted, centered at different locations across the workspace. The results obtained from the investigation are shown in Table~\ref{Line_circle_resistnace}. Both length and resistance of lines and circles showed a high repeatability, indicating a strong reliability on the pump-nozzle ensemble during the drawing procedure.

\begin{table}[htbp]
\caption{Drawing Reliability}
\begin{center}
\begin{tabular}{|c|c|c|}
\hline
Shape Type & Resistance ($\Omega$) & Length ($cm$) \\ \hline
Line & 20$\pm$0.7 & 10$\pm$0.2 \\ \hline
Circle & 16$\pm$1.2 & 10$\pm$0.6  \\ \hline
\end{tabular}
\label{Line_circle_resistnace}
\end{center}
\end{table}


\section{METHODS}

\subsection{Mixed-Variable Bayesian Optimization}

Bayesian Optimisation (BO) has shown great success in optimizing expensive black-box functions \cite{ru2018fast}, \cite{NIPS2012_4522},  \cite{hernandezlobato}, \cite{7352306}, \cite{frazier2018tutorial}, \cite{alvi2019asynchronous}, which is ideal for robotic applications where each experiment is expensive to evaluate. In this work, however, shape types are encoded as categorical variables while shape locations are continuous, which defeats the common assumption that the BO acquisition function is differential over the input space. Various approaches have been proposed to handle mixed-type (i.e. categorical and continuous) inputs. The simplest method is to use one-hot encoding \cite{gpyopt2016} on the categorical space, which transforms categorical values to continuous ones on which standard BO can perform. However, one-hot encoding significantly increases the dimensions of the search space, making the continuous optimization of the acquisition function much harder \cite{pmlr-v70-rana17a}, which is not feasible for context where there are only limited budget of trials.

This work adopts a state-of-the-art BO approach for optimizing mixed-type problems called Continuous and Categorical Bayesian Optimization (CoCaBo) \cite{ru2020bayesian}. 
The algorithm first builds a multi-armed bandit (MAB) system to select promising categorical values and then applies BO on continuous variables. Authors of CoCaBo chose EXP3 \cite{doi:10.1137/S0097539701398375}  for the MAB algorithm, which is a standard solution for adversarial MAB \cite{Allesiardo:2017aa} where the reward distribution is affected by an adversarial agent.
We have chosen CoCaBo for two reasons: 1) the EXP3 algorithm enables fast selection of promising categories, significantly reducing the number of iterations.
2) CoCaBo shares information across different input types through a special kernel that efficiently leverages all available data. All these features provide advantages for efficient learning which is important for robotic applications where a limited budget of experiments are conducted.

\begin{figure}[t]
\centerline{\includegraphics[width=0.48\textwidth]{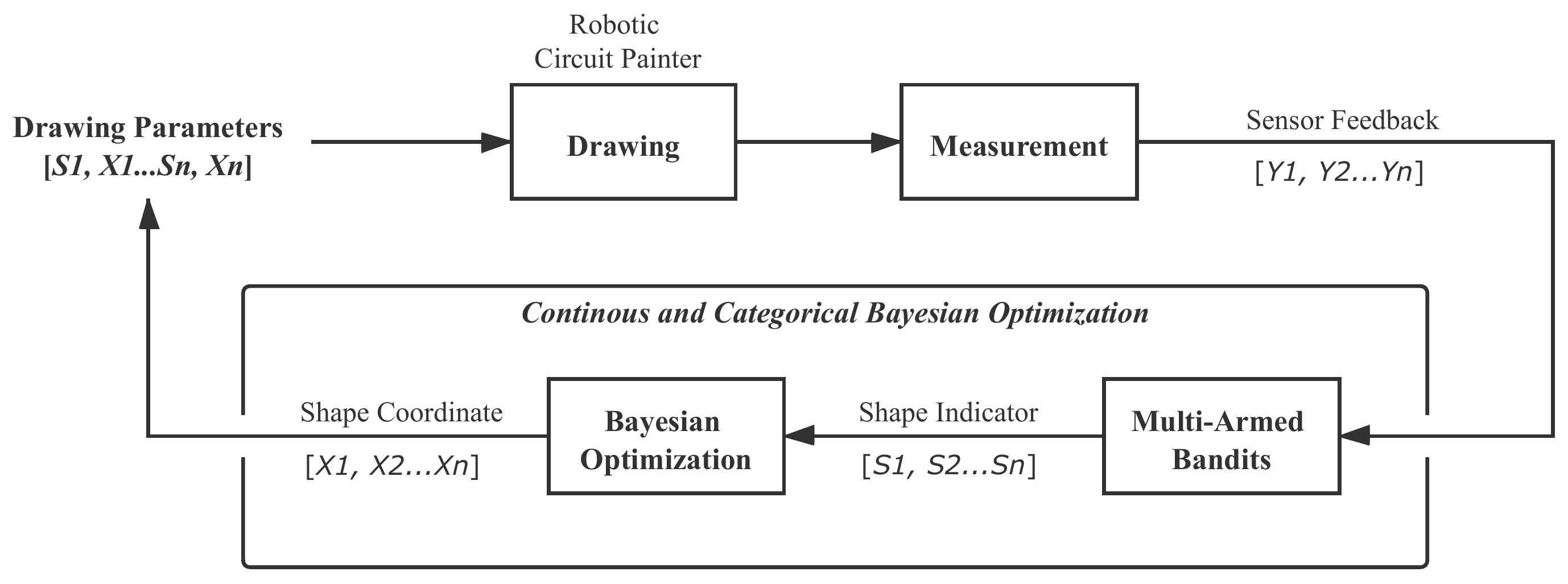}}
\caption{The Mixed-Variable Bayesian Optimization approach we investigate in this paper for circuit optimization.}
\label{CoCaBo_explain}
\end{figure}

\begin{figure}[t]
\centerline{\includegraphics[width=0.48\textwidth]{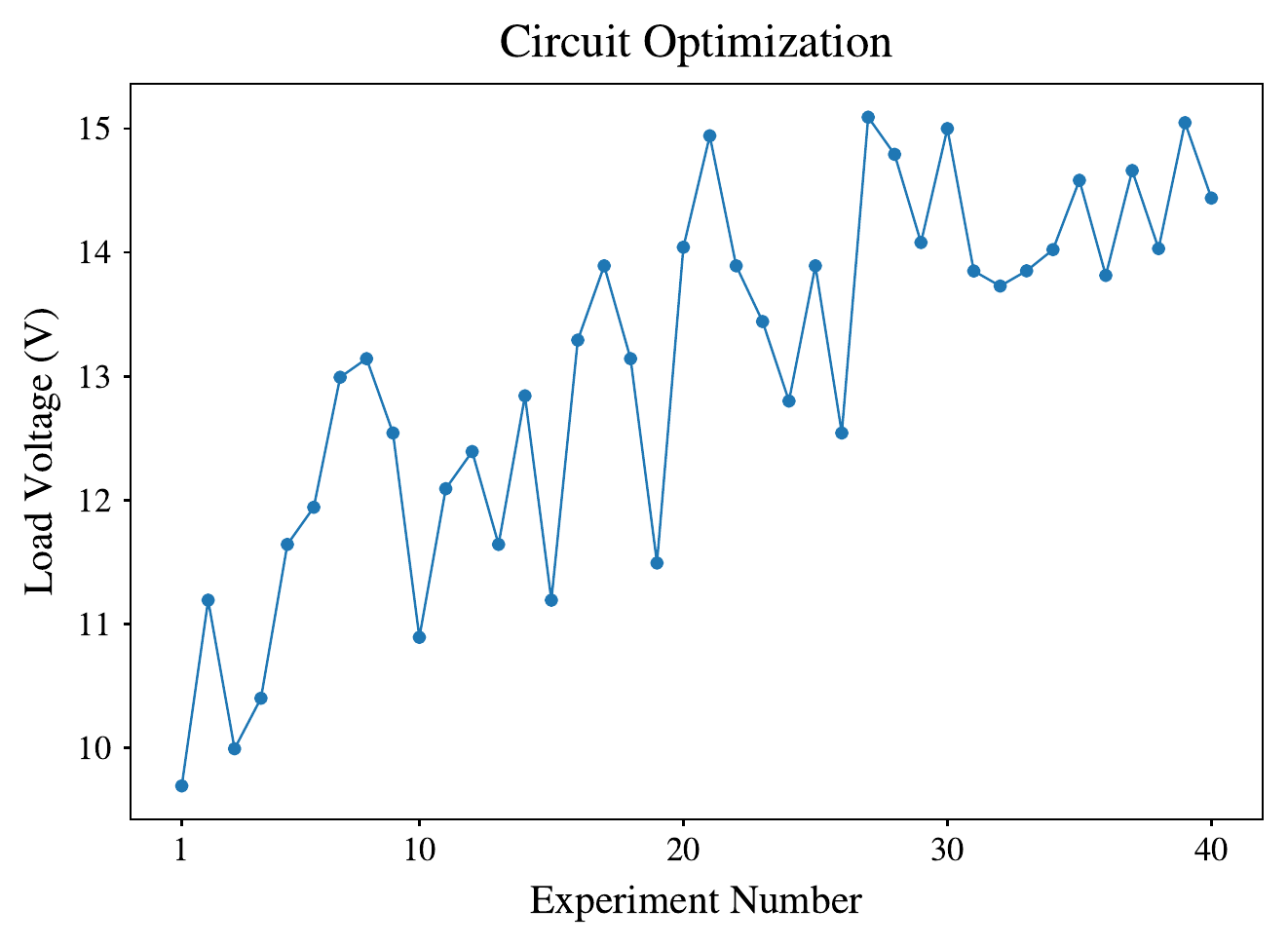}}
\caption{An incremental improvement in the voltage is achieved as the robot learns to choose circles over lines to reduce the overall resistance of the connection.}
\label{Exp1_voltageResult}
\end{figure}

\subsection{Reward Normalization}

The EXP3 algorithm is adopted as the MAB algorithm of CoCaBo for its rapid speed of selecting bandit candidates with a higher potential reward as the weights of arms in EXP3 change exponentially. However, only the weights of selected arms are updated, with all possible rewards being large and positive, and hence it is likely that the probabilities of good arms would grow too aggressively resulting in an insufficient exploration of the variable domain. In this work, if the load voltage ( $\approx10~V$ ) is directly used as the reward for optimization, the probability of any observed arm will increase aggressively, ignoring the overall performance. This becomes more problematic as the number of choices for each categorical variable is small. 
A common way to address this is to normalize the reward \cite{10.5555/3312046}. A set of initial test points is first selected to acquire reward samples, and the mean and standard deviation of the samples are calculated. The reward is then normalized by subtracting the mean and dividing by the standard deviation.

\subsection{Circuit Drawing and Optimization}

We modified the source code from the authors' GitHub repository (https://github.com/rubinxin/CoCaBO\_code) to perform the optimization process. 
For BO, the Gaussian Process (GP) employed a Matern kernel ($nu$ = $2.5$) and the exploration parameter ($k$) of the acquisition function based on the GP Upper Confidence Bound was set to be $2$.
As a demonstration of the effect of reward normalization, 10 trials were performed without reward normalization and the probabilities of circle category for $s_1-s_5$ were recorded. 
We constrained the experiment in 40 iterations with reward normalization to simulate the total operation budget the robot can perform before running out of energy. Five sets of initial parameters were randomly selected and the load voltage were observed, of which the mean and standard deviation were used for reward normalization.  

\subsection{Circuit Drawing and Optimization with Obstacle Avoidance}

In order to investigate the performance of the robot in uncertain environments, an obstacle was placed in the middle of the workspace. The obstacle is a metal bar connected to a small resistance ($5\Omega$) which significantly decreases the voltage on the robotic load. The location of the obstacle is designed that the circuit connects it when either $s_2$ or $s_4$ is too closed to the center (i.e., $x_1\geq0$ or $x_3\leq0$) or $s_3$ is a circle. 
Considering a wider range of rewards, we increased initial sets of parameters to 10 and in total 30 iterations were performed.


\section{RESULTS}

\subsection{Circuit Optimization}

The circuit optimization aims to minimize the resistance of the circuit connection, as this lower resistance directly translates as an improvement of the output voltage of the circuit. The observed voltage after 40 iterations is shown in Fig.~\ref{Exp1_voltageResult}. As the control parameters iterate through the experiment, the load voltage increases incrementally, indicating a loss reduction on the connection. 

\begin{figure}[t]
\centerline{\includegraphics[width=0.35\textwidth]{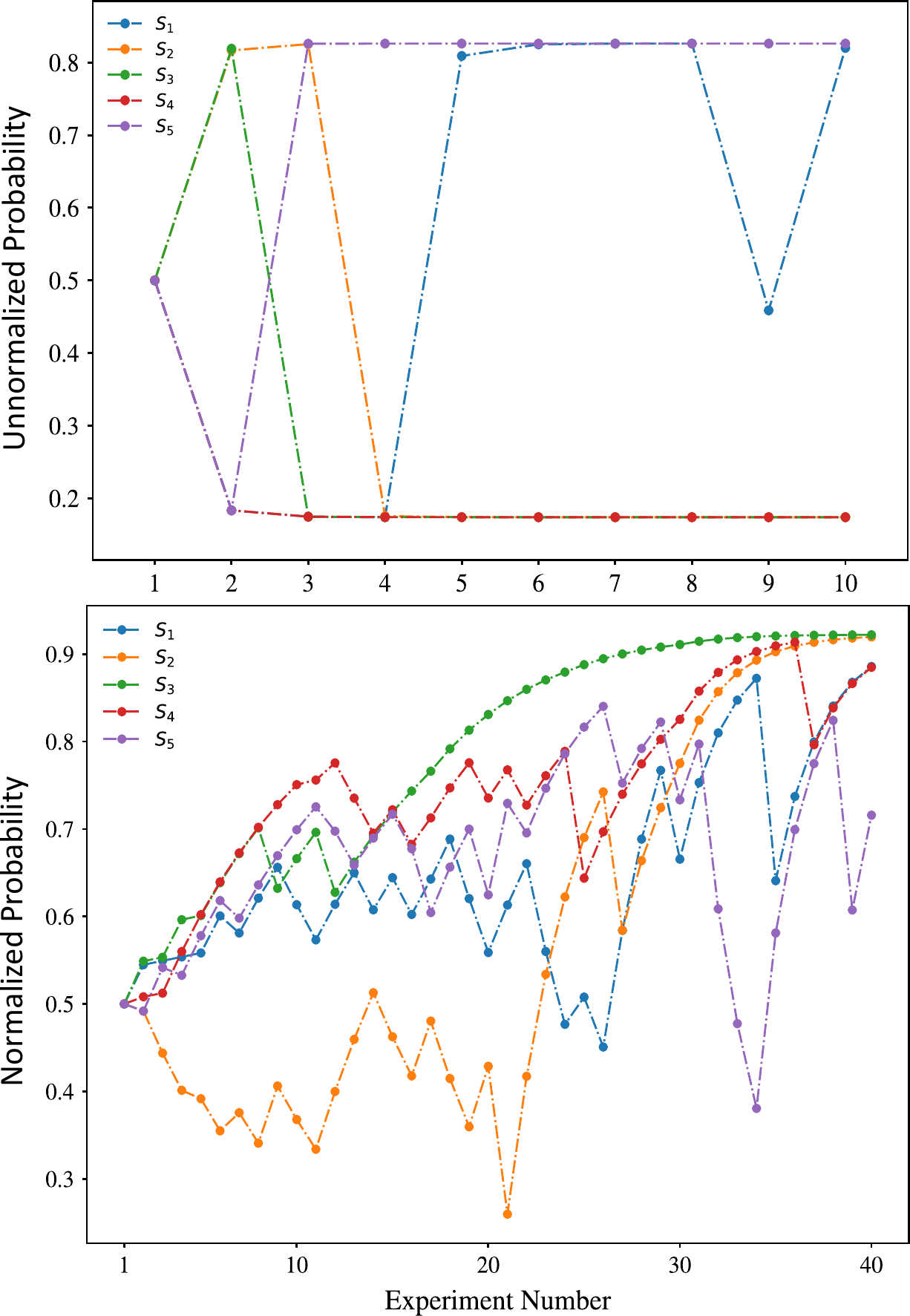}}
\caption{Without reward normalization, weights of chosen categories oscillate aggressively between 0.8 and 0.2, forcing us to end the experiment in the first ten iterations. By contrast, the implementation of reward normalization achieves smoother growth in promising categories, which is more likely to find a global optima. This allows experiments to start from 0.5 and gradually converge to higher results.}
\label{Exp1_circleProb}
\end{figure}

The advantage of applying reward normalization to the voltage is the prevention of aggressive growth in the probabilities of promising categories, which encourages exploration and enhances the likelihood of finding global optima. This can be seen in Fig.~\ref{Exp1_circleProb}. With proper reward normalization, the robot learns to choose circles over lines to reward itself with a high voltage.

\begin{figure}[t]
\centerline{\includegraphics[width=0.48\textwidth]{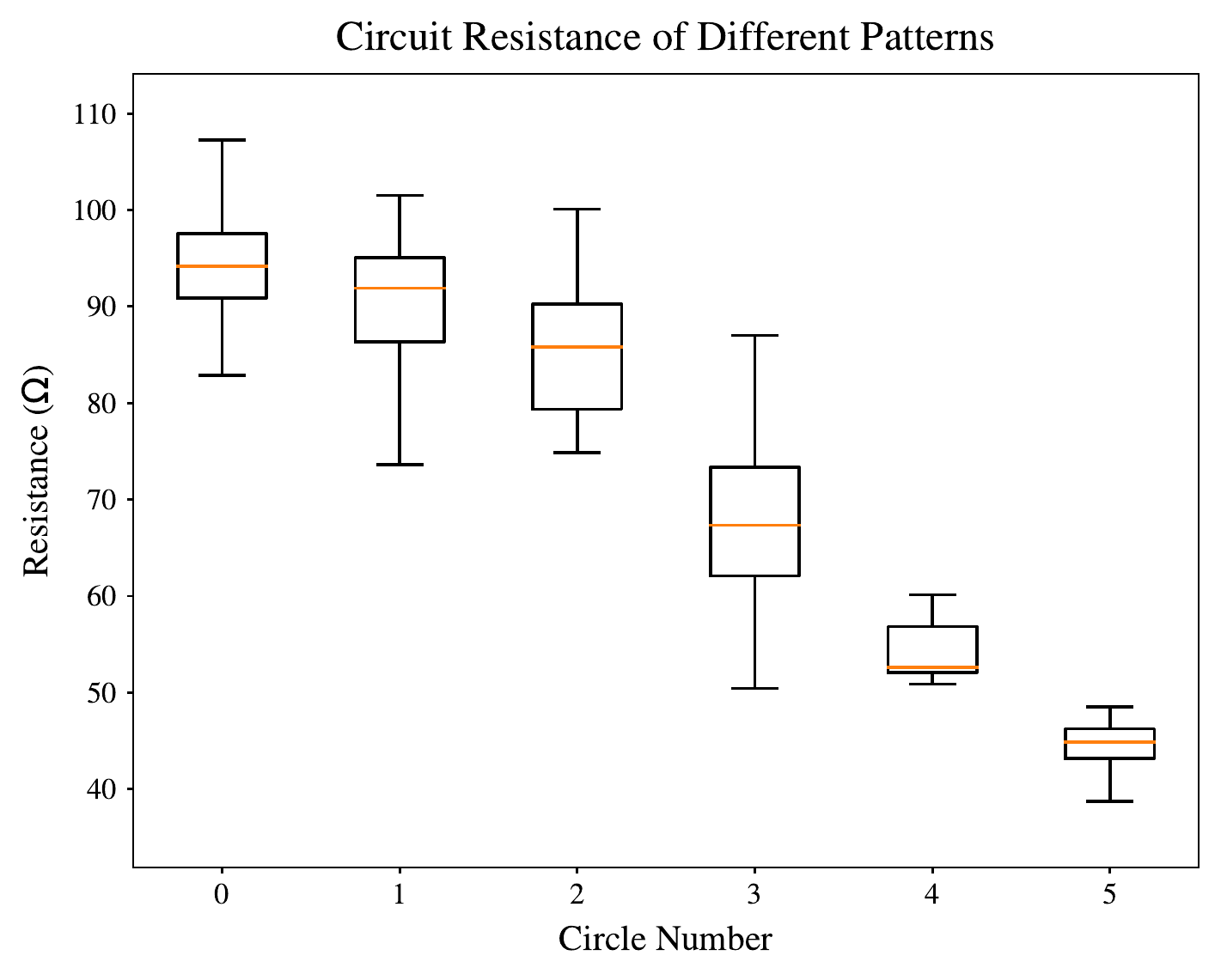}}
\caption{The overall resistance of connections decreases as there are more circles in the patterns. Significant reduction in resistance happens with 3 or more circles as more bridge circuits are created when circles intersect with each other.}
\label{CircleNum}
\end{figure}

\begin{figure}[t]
\centerline{\includegraphics[width=0.48\textwidth]{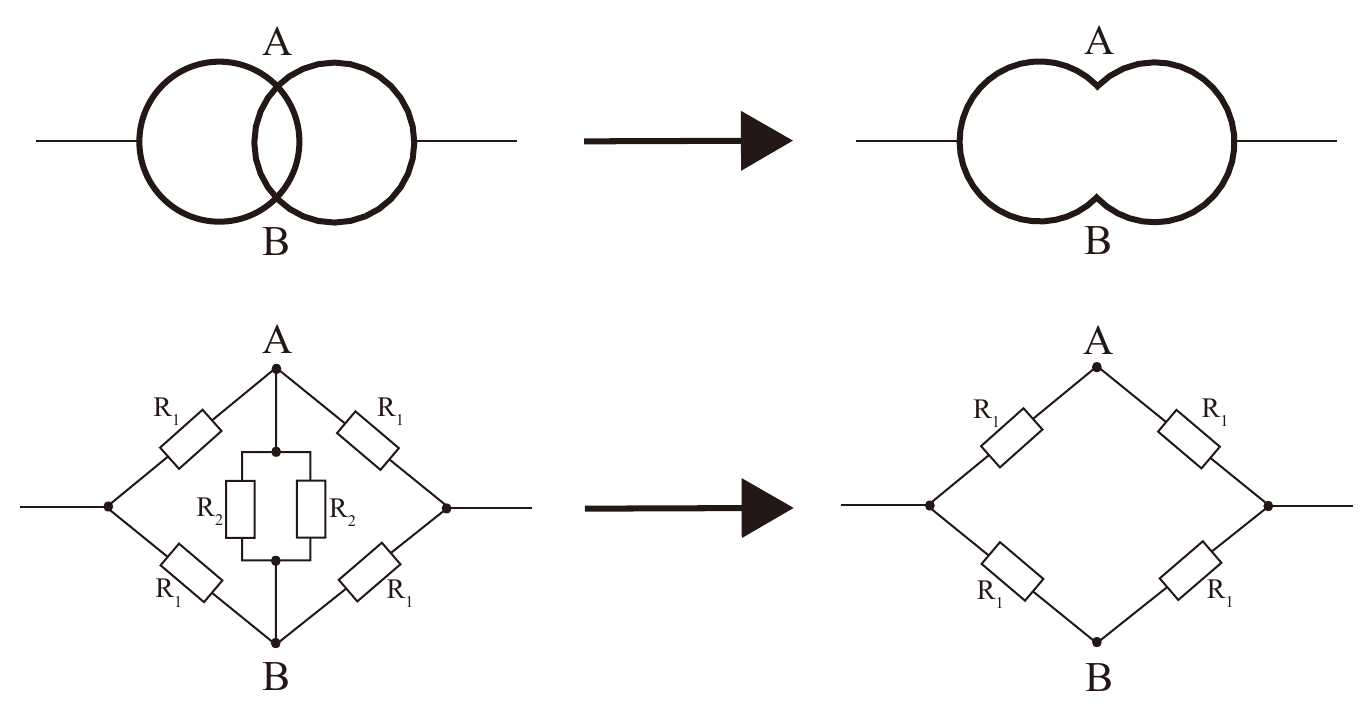}}
\caption{The circle-circle intersection creates a bridge circuit. Due to the symmetry, node A and node B theoretically have the same electric potential that no current will pass through $R_2$. The bridge circuit can then be simplified as a parallel circuit of four $R_1$, which causes a significant reduction in resistance.}
\label{equivalent circuit}
\end{figure}

The relationship between resistance of circuits and circle numbers is shown in Fig.~\ref{CircleNum}. We can see that moderate improvements are achieved by having one or two circles in the circuit patterns as circles have smaller resistance than lines. Interestingly, we observe significant improvements in the mean resistance when the circle number increases from two to five. This can be explained by the formation of bridge circuits as the result of circle-circle intersection. We further explain this with an equivalent circuit in Fig.~\ref{equivalent circuit}. Note that patterns with three circles present high variance in the resistance because the number of bridge circuits can change from zero to two, depending on the order of circles. Having more circles is more likely to create bridge circuits, which significantly reduces the overall resistance of the connection. Compared to the original connection with only lines, the resistance is halved after the circuit optimization.

\subsection{Circuit Optimization with Obstacle Avoidance}

\begin{figure}[t]
\centerline{\includegraphics[width=0.48\textwidth]{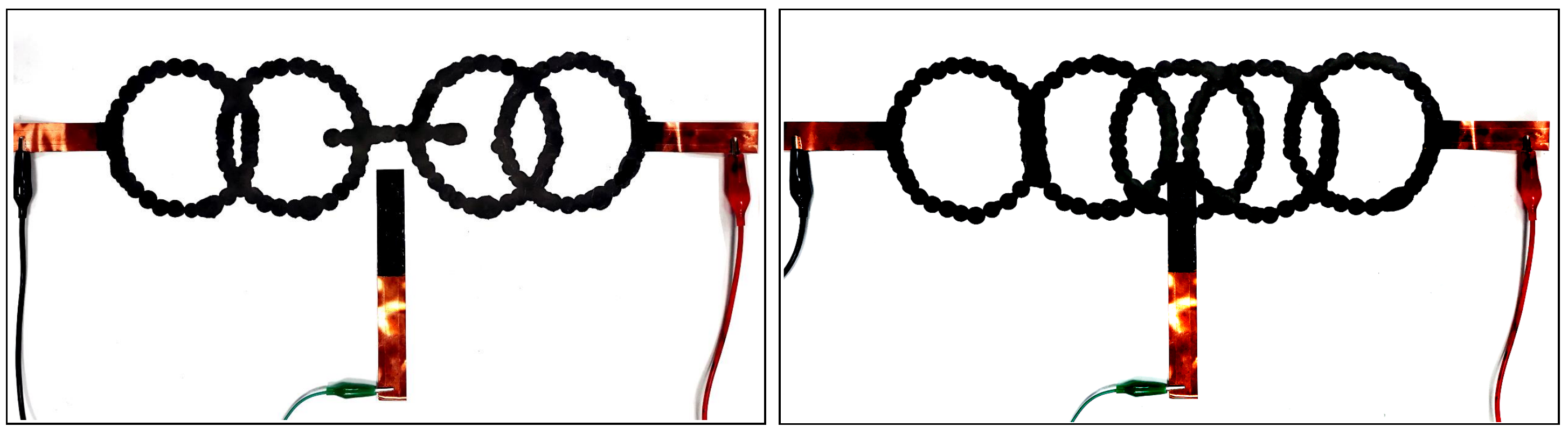}}
\caption{Examples of circuit patterns drawn during optimization. The left pattern has a high voltage as there are 4 circles and does not connect to the obstacle. The right one touches the obstacle, and thus shows a low voltage despite the fact that it has 5 circles.}
\label{Pattern}
\end{figure}

The purpose of introducing an obstacle in the path is to challenge the robot with a competitive and uncertain environment. The obstacle is linked to a small resistance. Once the circuit connects the obstacle, the voltage of the robotic load will decrease significantly.
As we have known from the results above, the load voltage is mostly affected by the number of circles. The optimal patterns in this setting are likely to have as many circles as possible while avoiding touching the obstacle. Fig.~\ref{Pattern} demonstrates two patterns drawn during the optimization process. 
In order to achieve optimal load voltage, the best strategy is to draw line for $s_3$ and draw circles for $s_2$ and $s_4$ while keeping $x_1$ and $x_3$ far away from the obstacle. This strategy requires shared information between categorical and continuous domains. 

\begin{figure}[t]
\centerline{\includegraphics[width=0.45\textwidth]{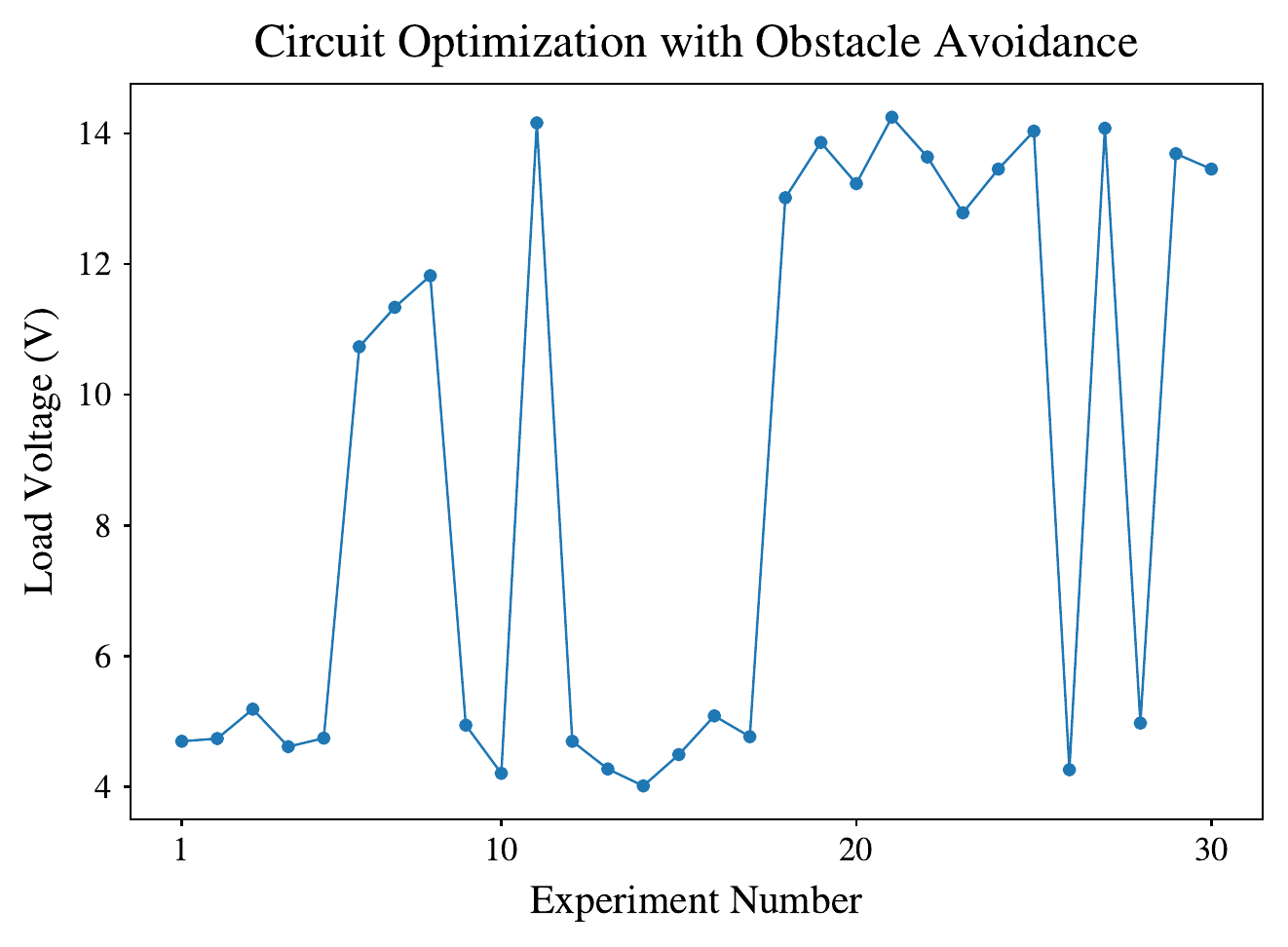}}
\caption{After introducing the obstacle, the load voltage experiences high-level of fluctuation at the beginning of the experiments but generally stabilized with a sequence of high load voltage.}
\label{Exp2_voltageResult}
\end{figure}


\begin{figure}[t]
\centerline{\includegraphics[width=0.48\textwidth]{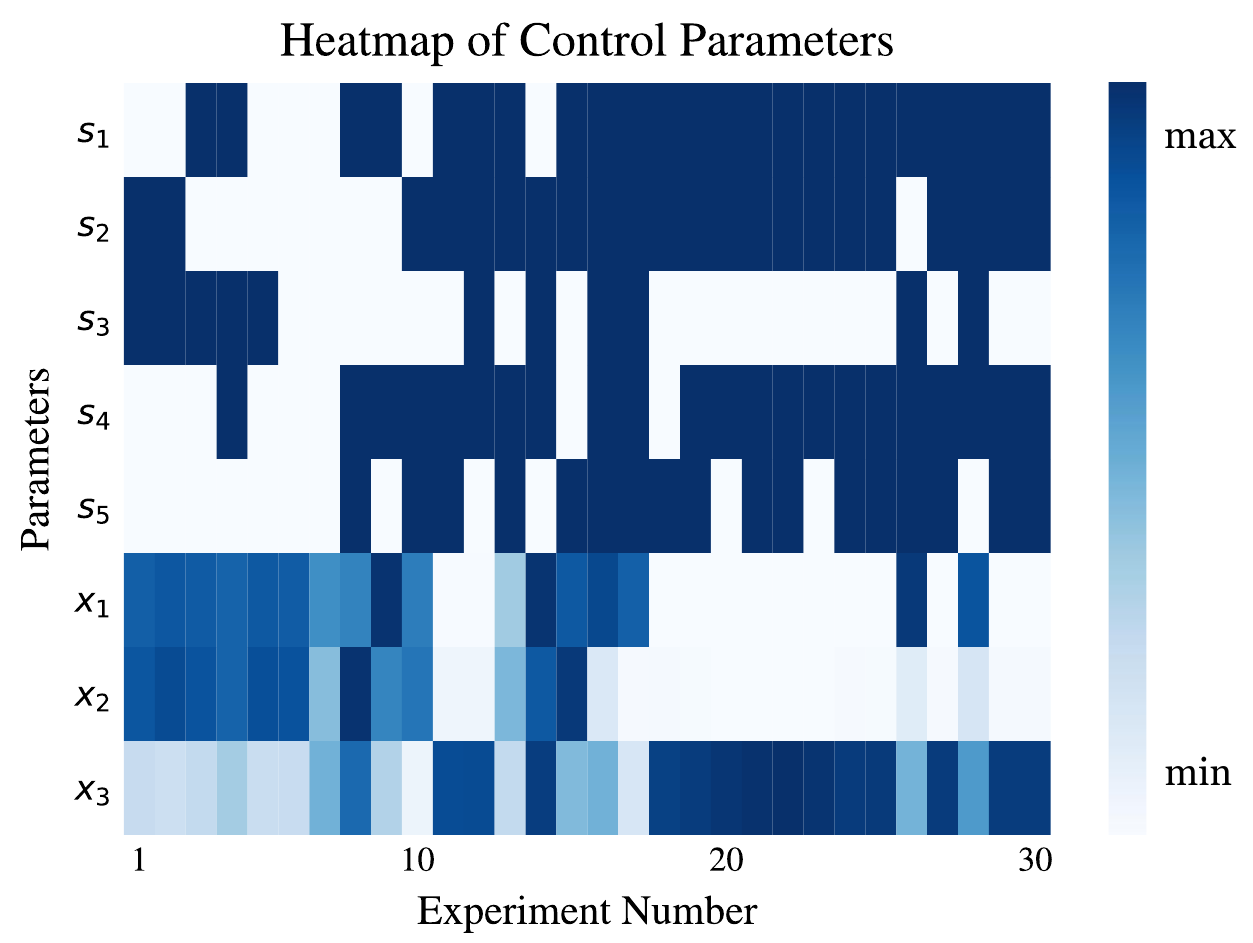}}
\caption{For $s_1-s_5$, dark blue denotes circles and white denotes lines. The obstacle can be avoided if $s_1$, $s_2$ and $s_3$ are all lines or $s_1$ and $s_3$ are circles centered away from the middle of the workspace.
When $s_2$ and $s_4$ are both circles, the robot learns to constrain $x_1$ and $x_3$ in a small search area to prevent connecting the obstacle.}
\label{Heat_map}
\end{figure}


The observed load voltage of 30 iterations is shown in Fig.~\ref{Exp2_voltageResult}. It is clear in the figure that the load voltage experiences severe fluctuation during the experiment compared to the situation without an obstacle. The load voltage remains low at the beginning as $s_3$ is a circle and therefore connects the obstacle. However, the robot learns to change $s_3$ to circle after a few iterations to avoid the obstacle, as shown in Fig.~\ref{Heat_map}. The most interesting aspect of Fig.~\ref{Heat_map} happens around iteration 20. As $s_2$ and $s_4$ are chosen to be circles, the values of $x_1$ and $x_3$ are constrained in a very small area that allow the center of $s_2$, $s_4$ away from the obstacle, which results in high load voltage. This can be explained by the successful coupling between the continuous and categorical domains, which is through a special combination of kernels of different variables designed in CoCaBo.


\section{CONCLUSIONS}
This paper presents a novel approach for robots to access energy to survive in energy-limited and uncertain environments, through self-drawn electrical connections using graphene-based conductive ink. Two tasks were performed where a robot needs to replace series connections with parallel ones while avoiding obstacles to receive maximum power from the power supply. With properly designed optimization routines, the robot can succeed in these tasks within a small number of trials. This study indicates that robots can survive and be self-sufficient by combining self-drawn electric circuits and Bayesian Optimization. As robots become more present in our society and even reach other planets, maximizing their capacity to keep themselves online is crucial to increase the odds of success.

Apart from energy, other resources are also crucial for robotic survival. Keeping the usage of material (e.g., conductive ink) in the optimization routine is a further direction to explore. For example, circles outperform lines in reducing resistance of the connection but cost more ink. The robot should consider the trade-off between the usage of resources and the improvement in received energy. Currently, we only implement circuit drawing on flat surfaces.  Applying circuit drawing in 3-dimensional space would be another interesting further direction.


\addtolength{\textheight}{-12cm}   








\bibliographystyle{ieeetr}
\bibliography{egbib}

\end{document}